\pdfoutput=1

\documentclass[11pt]{article}

\usepackage[final]{acl}

\usepackage{times}
\usepackage{latexsym}

\usepackage[T1]{fontenc}

\usepackage[utf8]{inputenc}

\usepackage{microtype}

\usepackage{inconsolata}

\usepackage{graphicx}
\usepackage{xcolor}
\usepackage{subcaption}
\usepackage{arydshln}
\usepackage{pifont}
\usepackage{adjustbox}
\usepackage{booktabs}
\usepackage{multirow}
\usepackage{colortbl}
\usepackage{amssymb}
\usepackage{cleveref}
\usepackage{array}
\usepackage{tcolorbox}

\crefformat{section}{\S#2#1#3}
\crefformat{subsection}{\S#2#1#3}
\crefformat{subsubsection}{\S#2#1#3}

\title{Evaluating Visual and Cultural Interpretation:\\The \texttt{K-Viscuit} Benchmark with Human-VLM Collaboration}

\author{ChaeHun Park\footnotemark[1]$^1$  \hspace{0.3cm} Yujin Baek\thanks{\hspace{0.2cm} Equal contribution}$^1$\hspace{0.3cm}  Jaeseok Kim$^2$ \\
\hspace{0.3cm}     \textbf{Yu-Jung Heo}$^2$\hspace{0.3cm}    \textbf{Du-Seong Chang}$^3$ \hspace{0.3cm} \textbf{Jaegul Choo}$^1$\\
$^1$ KAIST AI \hspace{0.3cm} $^2$ KT Corporation \hspace{0.3cm} $^3$ Sogang Univ.\\
\hspace{0cm}\texttt{\{ddehun,yujinbaek,jchoo\}@kaist.ac.kr} \\
\hspace{0cm}\texttt{\{jaeseok.kim,yj.heo\}@kt.com} 
\hspace{0.4cm}\texttt{duseong.chang@gmail.com} \\
 }

\begin{document}
\maketitle
\begin{abstract}
To create culturally inclusive vision-language models (VLMs), developing a benchmark that tests their ability to address culturally relevant questions is essential. Existing approaches typically rely on human annotators, making the process labor-intensive and creating a cognitive burden in generating diverse questions. To address this, we propose a semi-automated framework for constructing cultural VLM benchmarks, specifically targeting multiple-choice QA. This framework combines human-VLM collaboration, where VLMs generate questions based on guidelines, a small set of annotated examples, and relevant knowledge, followed by a verification process by native speakers. We demonstrate the effectiveness of this framework through the creation of \texttt{K-Viscuit}, a dataset focused on Korean culture. Our experiments on this dataset reveal that open-source models lag behind proprietary ones in understanding Korean culture, highlighting key areas for improvement. We also present a series of further analyses, including human evaluation, augmenting VLMs with external knowledge, and the evaluation beyond multiple-choice QA. Our dataset is available at \url{https://huggingface.co/datasets/ddehun/k-viscuit}.
\end{abstract}

\section{Introduction}
\label{sec:introduction}


Recent advances in vision-language models (VLMs) have demonstrated remarkable capabilities in tasks ranging from image captioning to visual question answering. However, these models are predominantly trained on Western-centric datasets \citep{lin2014microsoft, young2014image, antol2015vqa}, leading to substantial performance disparities when applied to non-Western contexts \citep{marvl, yin2021broaden, yin2023givl, romero2024cvqa}. This cultural bias is particularly problematic as visual interpretation often depends heavily on cultural context, necessitating the development of more culturally aware VLMs.

\begin{figure*}[t!]
\centering
\includegraphics[width=0.95\textwidth]{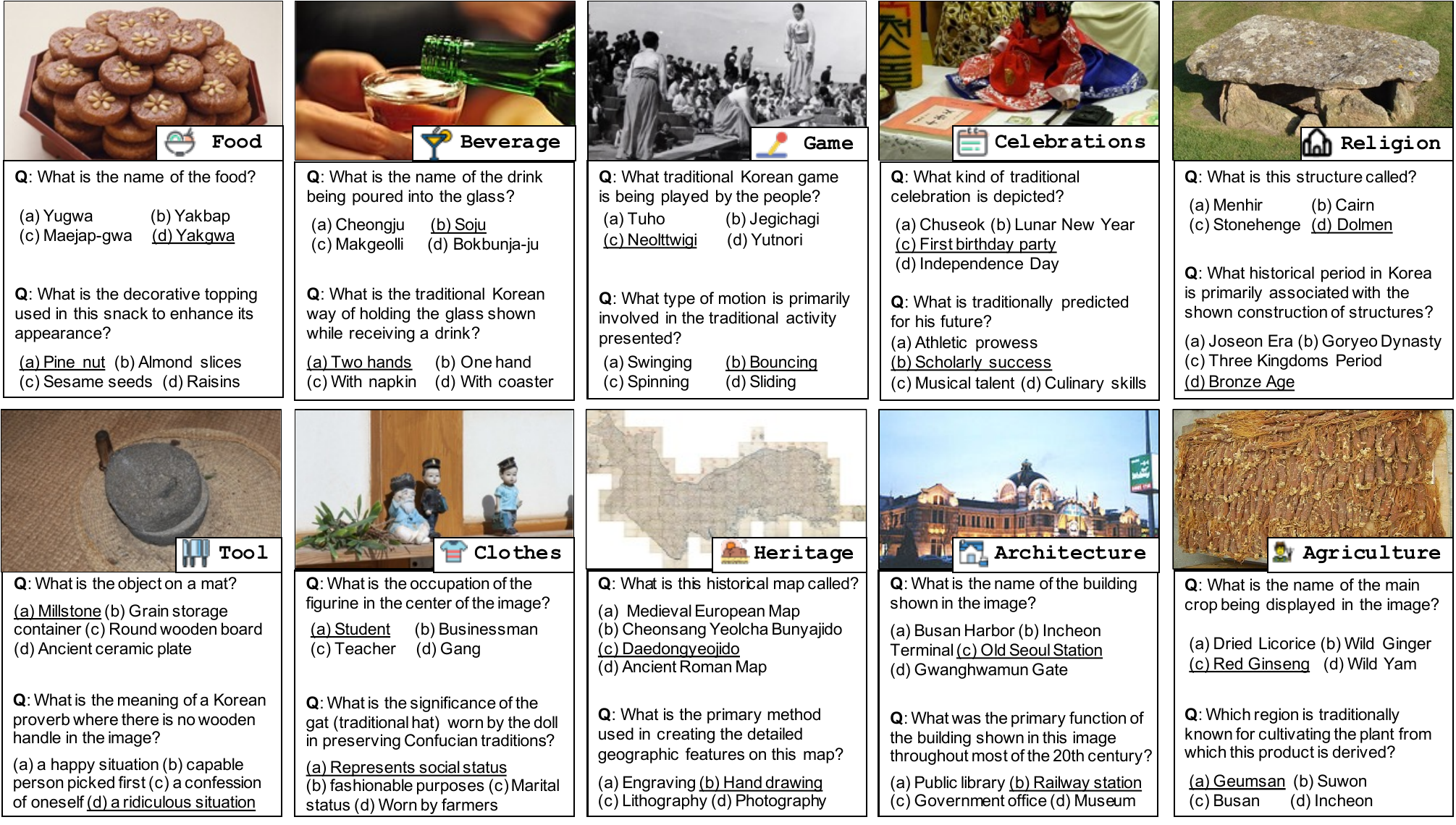}
\caption{\textbf{Dataset Examples}. We present an image and two questions of different types for each concept category.}
\label{fig:data_example}
\end{figure*}



Several benchmarks have been proposed to evaluate cultural understanding in VLMs \citep{marvl, yin2021broaden, romero2024cvqa, nayak2024benchmarking,bhatia-etal-2024-local}. These approaches primarily rely on manual question generation, which, while valuable, faces certain practical challenges. The manual process can be time-consuming and resource-intensive when scaling to new cultural contexts. Additionally, as noted in cognitive science research \citep{ramos2020cognitive}, humans may experience cognitive fixation, potentially limiting the diversity of generated questions. These practical considerations motivate the need for more efficient benchmark construction approaches.

\begin{figure*}[t!]
\centering
\includegraphics[width=0.9\linewidth]{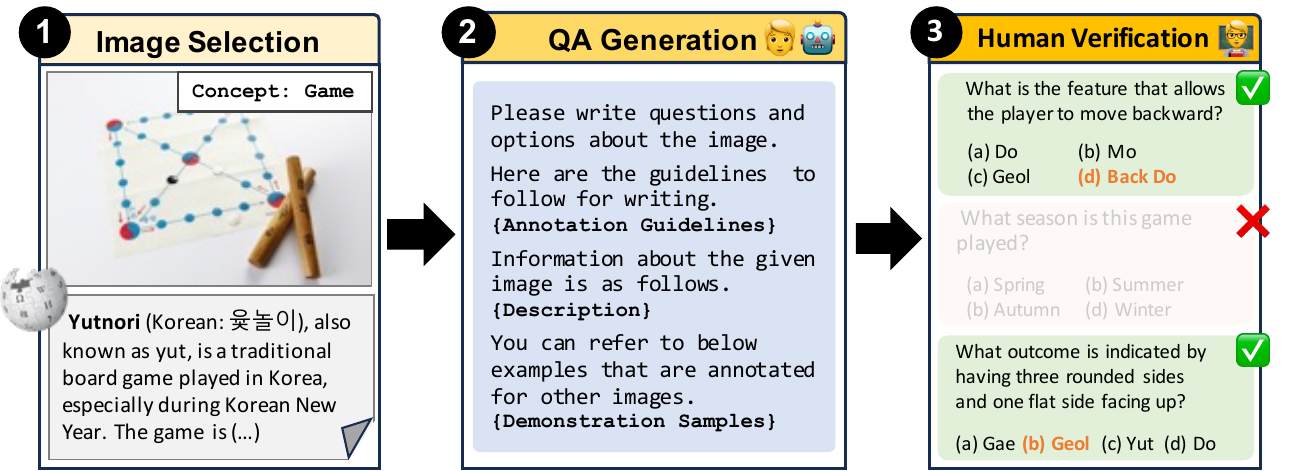}

\caption{\textbf{Framework Overview.} We begin with image selection, where concept-relevant images (e.g., the traditional Korean game Yutnori) are chosen and briefly described. In the QA generation stage, a VLM is guided by annotation guidelines and demonstration samples to draft question–option pairs. Finally, human verification ensures annotator review, correction, and filtering of the generated samples.}
\label{fig:pipeline_overview}
\end{figure*}

Inspired by recent successes in human-LLM collaborative data generation~\citep{liu2022wanli,bartolo2022models, kamalloo2023hagrid}, we propose a semi-automated framework for constructing cultural VLM benchmarks that enhances both the efficiency and diversity of culture-relevant visual question and answer generation (see Fig.~\ref{fig:pipeline_overview}). Our framework incorporates human-VLM collaboration, where the VLM generates and recommends questions and answers based on carefully crafted guidelines, a small set of human-annotated examples, and image-specific knowledge. Native speakers then verify these recommended questions to ensure quality and cultural relevance.


Using this framework, we develop \texttt{K-Viscuit} (Korean Visual and Cultural Interpretation Test), a benchmark dataset tailored to Korean culture. 
Note that the framework is adaptable and can be similarly applied to other cultures.
\texttt{K-Viscuit} features two distinct evaluation types: visual recognition and visual reasoning. Also, the benchmark employs carefully designed multiple-choice questions with highly similar distractors to prevent models from exploiting superficial patterns.

Our evaluation with \texttt{K-Viscuit} reveals a notable performance gap between open-source and proprietary VLMs in understanding Korean culture. We provide insights into the current limitations and potential improvements in VLMs' cultural understanding capabilities through detailed analyses, which include human evaluation, external knowledge integration, and extended evaluation beyond a multi-choice question-answering setup.

Our contributions are summarized as follows:
\begin{itemize}
\item We propose a semi-automated framework for constructing benchmarks to evaluate the cultural understanding capabilities of VLMs.
\item We develop \texttt{K-Viscuit}, a Korean culture-focused VQA benchmark using our proposed framework.
\item We present comprehensive experimental results and analyses of both open-source and proprietary VLMs evaluated on \texttt{K-Viscuit}.
\end{itemize}

\section{Related Work}
\label{sec:related work}

Recent research has made significant strides in developing benchmarks to assess AI models' cultural understanding capabilities.
These efforts are particularly important as many existing models, including VLMs and Large Language Models (LLMs), are trained predominantly on Western-centric datasets~\citep{young2014image, lin2014microsoft,antol2015vqa}, limiting their effectiveness in non-Western contexts.
For LLMs, several notable cultural benchmarks have emerged: \citet{kim2024click} introduced \texttt{CLIcK}, a benchmark for evaluating Korean language models' cultural knowledge through carefully designed QA pairs, while \citet{wibowo2023copalID} developed \texttt{COPAL-ID} to capture Indonesian cultural nuances in text-based commonsense reasoning.

In the multimodal domain, VLMs require consideration of both visual and textual inputs to effectively reflect cultural contexts.
\citet{marvl} addressed this challenge with \texttt{MaRVL}, a multilingual visually-grounded reasoning dataset spanning five languages and cultures, demonstrating the importance of cross-cultural visual-linguistic understanding.
Building on this foundation, \citet{yin2021broaden} introduced \texttt{GD-VCR} to evaluate geographical and cultural aspects of visual commonsense reasoning, while \citet{romero2024cvqa} developed \texttt{CVQA}, a comprehensive multilingual VQA benchmark for assessing VLMs across diverse cultural contexts.

While these cultural benchmarks have provided valuable insights, their reliance on manual annotation can constrain both the diversity and efficiency of dataset creation~\citep{marvl,yin2021broaden,ramaswamy2024geode,romero2024cvqa}.
Recent work has demonstrated the potential of AI-assisted dataset generation when combined with human expertise.
\citet{liu2022wanli} successfully applied this approach to natural language inference tasks, and similar strategies have been widely adopted in creating language-only datasets~\citep{taori2023alpaca, kim2023aligning, dubois2024alpacafarm, kim2024click}.
Our work extends this paradigm to multimodal cultural benchmarks, leveraging AI models to enhance dataset diversity while maintaining high quality through human verification.

\section{Data Construction Framework}
\label{sec:dataset}
We present our human-AI collaborative framework for constructing datasets to evaluate VLMs' understanding of specific cultural domains.
In this work, we focus on Korean culture as our target domain.
First, we provide an overview (\cref{subsec:data_framework}) and implementation details.
Then, we present the analysis of our resulting dataset, \texttt{K-Viscuit} (\textbf{K}orean \textbf{Vis}ual and \textbf{Cu}ltural \textbf{I}nterpretation \textbf{T}est)~(\cref{subsec:data_analysis}).

\subsection{Framework Overview}
\label{subsec:data_framework}
Our framework is designed to create a multiple-choice visual question answering (VQA) task, where each evaluation sample consists of an image, a question, and four options with one correct answer.
Native Korean speakers participate in the dataset construction process, while a powerful proprietary VLM is employed to mitigate unintended human biases, such as cognitive fixation~\citep{ramos2020cognitive}, and streamline the annotation process.
The generated samples cover various aspects of the target culture derived from daily life and require multimodal reasoning to interpret both visual and textual information accurately.
The dataset construction consists of four stages: 1) concept selection, 2) image selection, 3) question and options annotation, and 4) human verification.
Fig.~\ref{fig:pipeline_overview} illustrates the overall framework.

\subsubsection{Concept Categorization}
Inspired by recent studies on multicultural evaluation datasets \citep{marvl,wibowo2023copalID,kim2024click}, we aim to assess knowledge of various concepts encountered in daily life by Korean natives.
While each concept should have some degree of universality, its manifestation often varies across cultures.
Following \citet{marvl}, we reference semantic concepts from the Intercontinental Dictionary Series (IDS) \citep{key_comrie_2015} to define our concept list.
Our dataset encompasses ten core concepts: \textsc{Food}, \textsc{Beverage}, \textsc{Game}, \textsc{Celebrations}, \textsc{Religion}, \textsc{Tool}, \textsc{Clothes}, \textsc{Heritage}, \textsc{Architecture}, and \textsc{Agriculture}. Further details are in Appendix \ref{sec:app:ids_mapping}.

\subsubsection{Image Selection}
Korean native annotators collected web images corresponding to the selected concepts.
To ensure diverse representation, we limited each specific object to appearing no more than twice within any single category.
Following \citet{marvl}, we selected only images depicting concepts that could physically exist in everyday life.
Annotators were encouraged to source diverse and suitable images from various web resources.\footnote{Although framed as image selection, this step also served as a concept-level filtering process. For instance, we excluded abstract or symbolic visuals (e.g., fireworks emojis for “celebration”) that lacked concrete visual referents.}
Wikimedia Commons\footnote{\url{https://commons.wikimedia.org/wiki}} served as the primary source, and only CC-licensed images were selected.

\subsubsection{Question Generation}
\noindent\textbf{{Question Type}}
Based on the selected images, we annotate questions in a multiple-choice QA format.
To comprehensively evaluate the understanding of Korean culture, we categorize questions into two types: Visual Recognition (\textsc{Type 1}) and Cultural Knowledge Application (\textsc{Type 2}).
\textsc{Type 1} questions assess basic visual information, such as object identification.
In contrast, \textsc{Type 2} questions require a deeper understanding of cultural context, reasoning, or multi-step inference based on the image.
For each image, we created one \textsc{Type 1} question and between one and four \textsc{Type 2} questions.
This categorization provides two key advantages:
First, \textsc{Type 1} questions assess a model's ability to recognize culturally embedded visual elements.
Second, \textsc{Type 2} questions evaluate cultural understanding beyond simple object recognition.

\begin{table}[t!]
\centering
\small
\begin{tabular}{lc}
\toprule[1.2pt]
\noalign{\vspace{-0.3ex}}
\# of samples &   657 \\
\,\, - \textsc{Type 1}/ \textsc{Type 2} & 237/420 \\
\# of unique images & 237 \\
\# of options & 2628 \\
\,\, \# of unique options & 2129 \\
\noalign{\vspace{0.1ex}}
\hline
\noalign{\vspace{0.1ex}}
Avg. question length & 13.5 \\
\,\, - \textsc{Type 1}/ \textsc{Type 2} & 10.1/15.5 \\
Avg. option length & 1.7 \\
\noalign{\vspace{-0.3ex}}
\bottomrule[1.2pt]
\end{tabular}
\caption{\textbf{Dataset statistics}. The length of questions and options denotes the number of words.}
\label{tab:data_stat}
\end{table}
\begin{figure}[t!]
\centering
\includegraphics[width=0.75\linewidth]{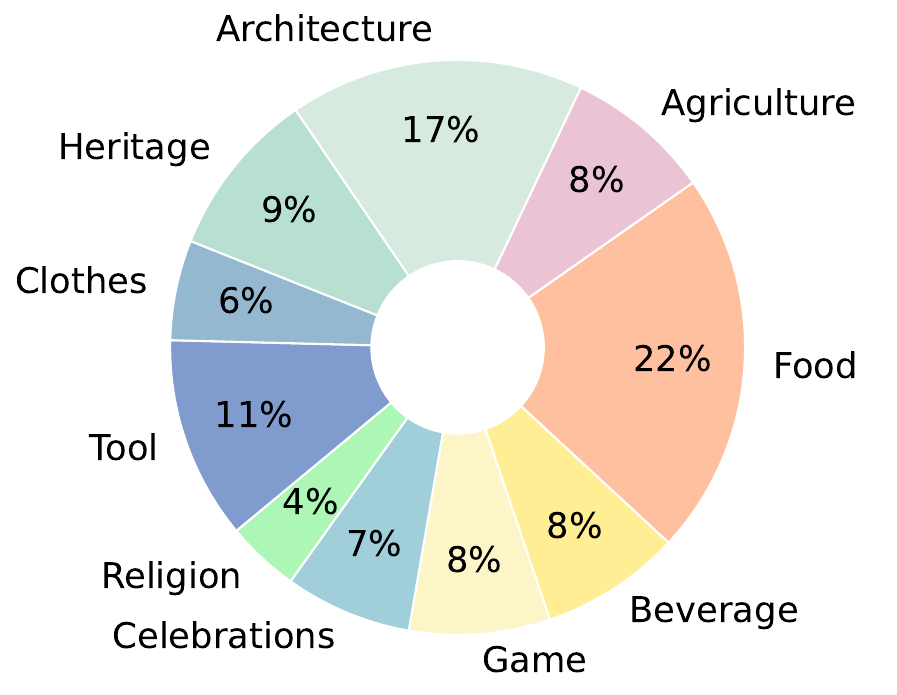}
\vspace{-0.2cm}
\caption{Concept distribution of our dataset.}
\label{fig:pie_chart}
\end{figure}

\noindent\textbf{{AI-assisted Question Annotation}}
We create questions and their options (one correct answer and three distractors) by leveraging a powerful proprietary VLM (GPT-4-Turbo).
For each concept category, human annotators first create exemplar questions and options for at least three images.
These manually annotated examples serve as demonstrations for the VLM to generate additional questions and options.

Specifically, the VLM receives: 1) the target image, 2) human-annotated demonstration examples, 3) detailed annotation guidelines, and 4) image-specific knowledge descriptions.
We include relevant contextual knowledge for each image to enhance question diversity and relevance, ensuring VLM-generated questions are grounded in real-world understanding.
Notably, following \citet{wang2023mcq}, our guidelines emphasize maintaining high similarity among all four multiple-choice options, a principle also reflected in human-annotated examples.
All information is provided to the VLM through natural language prompts, with distinct annotation guidelines for visual recognition and application questions.
The detailed prompts are presented in Appendix~\ref{sec:app_data_detail}.

\begin{figure}[t!]
\centering
\includegraphics[width=0.85\linewidth]{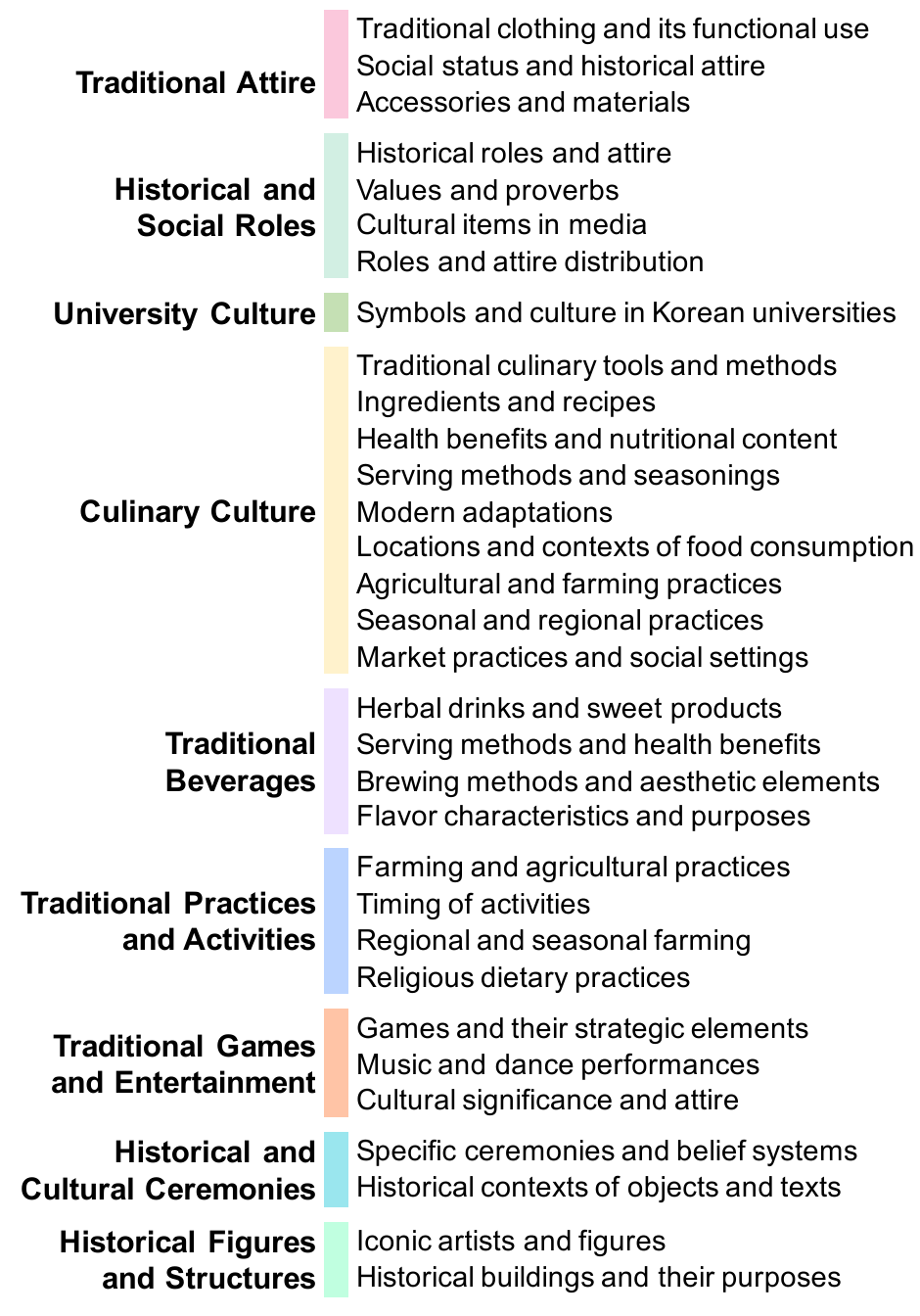}
\vspace{-0.2cm}
\caption{Required Cultural Knowledge.}
\label{fig:cultural_knowledge}
\end{figure}

It should be noted that all text in the dataset is written in English to isolate the evaluation of multicultural comprehension from multilingual aspects. However, as culture and language are not entirely orthogonal \citep{susan1996language,kramsch2014language}, completely separating them can be challenging. For instance, we observed that certain Korean terms lack exact English equivalents. In such cases, we romanize these Korean terms following standard transliteration rules.

\begin{table*}[t!]
\centering
\begin{adjustbox}{width=\textwidth}
\begin{tabular}{l|c|cccccccccc}
\toprule

\textbf{Model}&	\textbf{All}&	\textbf{Food}&	\textbf{Beverage}&	\textbf{Game}&	\textbf{Celeb.}&	\textbf{Religion}&	\textbf{Tool}&	\textbf{Clothes}&	\textbf{Heritage}	&\textbf{Arch.}	&\textbf{Agri.} \\ \midrule

InstructBLIP-7B~\citep{dai2024instructblip}          & 50.84                   & 40.85                    & 42.31                        & 38.46                    & 53.19                            & 40.74                        & 50.67                    & 62.16                       & 51.61                        & 60.55                            & 72.22                           \\
instructBLIP-13B~\citep{dai2024instructblip}        & 55.56                   & 45.77                    & 50.00                        & 46.15                    & 59.57                            & 55.56                        & 54.67                    & 64.86                       & 66.13                        & 60.55                            & 64.81                           \\
mPLUG-Owl2-7B~\citep{ye2023mplug}         & 48.25                   & 42.25                    & 42.31                        & 30.77                    & 63.83                            & 55.56                        & 48.00                    & 54.05                       & 45.16                        & 49.54                            & 66.67                           \\
LLaVA-1.6-7B~\citep{liu2024visual}       & 56.32                   & 43.66                    & 48.08                        & 40.38                    & 57.45                            & 51.85                        & 54.67                    & 67.57                       & 59.68                        & 72.48                            & 72.22                           \\
LLaVA-1.6-13B~\citep{liu2024visual}          & 57.08                   & 45.07                    & 53.85                        & 36.54                    & 68.09                            & 40.74                        & 53.33                    & 70.27                       & 66.13                        & 69.72                            & 70.37                           \\
InternLM-XC2-7B~\citep{dong2024internlm}        & 59.67                   & 50.70                    & 48.08                        & 40.38                    & 65.96                            & 55.56                        & 58.67                    & 64.86                       & 69.35                        & 69.72                            & 75.93                           \\
Molmo-7B-D \citep{deitke2024molmo}      	&61.04&	58.45&	71.15&	44.23&	68.09&	44.44&	61.33	&70.27&	56.45&	64.22&	68.52 \\
Idefics2-8B~\citep{2024Idefics2}       & 63.62                   & 51.41                    & 50.00                        & 50.00                    & 74.47                            & 66.67                        & 69.33                    & 75.68                       & 74.19                        & 73.39                            & 62.96                           \\
Llama-3.2-11B \citep{dubey2024llama} 	&68.04&	61.27&	65.38&	50.00&	72.34&	70.37	&72.00	&75.68&	72.58&	69.72&	81.48 \\

\midrule
Claude-3-opus~\citep{anthropic2024claude}       & 70.02                   & 62.68                    & 73.08                        & 59.62                    & 72.34                            & 77.78                        & 74.67                    & 78.38                       & 75.81                        & 67.89                            & 75.93                          \\
GPT-4-Turbo~\citep{gpt4}             & 80.82                   & 73.94                   & \underline{80.77}                        & \underline{78.85}                    & \underline{85.11}                            & \textbf{85.19}                        & 81.33                    & \underline{86.49}                       & \underline{85.48}                        & 79.82                            & \textbf{87.04}                         \\
Gemini-1.5-Pro~\citep{reid2024gemini}           & \underline{81.58}                   & \underline{80.28}                    & 78.85                        & 71.15                    & \underline{85.11}                            & 77.78 & \underline{82.67}& 83.78& 83.87                        & \underline{84.40}                            & 85.19                           \\

GPT-4o \citep{openai2024gpt4o}           & \textbf{89.50}  & \textbf{88.73}&	\textbf{82.69}&	\textbf{86.54}&	\textbf{95.74}&	\textbf{85.19}	&\textbf{90.67}	&\textbf{91.89}	&\textbf{91.94}&	\textbf{91.74}	&\textbf{87.04} \\

\bottomrule
\end{tabular}
\end{adjustbox}
\caption{\textbf{VLMs Evaluation Results.} \textit{Celeb.}, \textit{Arch.}, and \textit{Agri.} denote \textit{Celebration}, \textit{Architecture}, and \textit{Agriculture}. The highest and the second highest scores in each column are highlighted in bold and underlined.}
\label{tab:main_table}
\vspace{-0.3cm}
\end{table*}

\subsubsection{Human Verification}
In this step, we evaluate whether the generated QA pairs adhere not only to the detailed prompt guidelines (Tables \ref{tab:type1_prompt} and \ref{tab:type2_prompt}), but also to the intended cultural context. Our preliminary studies revealed that although VLMs often produced factually accurate outputs, some failed to capture the nuanced cultural elements we aimed to represent. 
Rather than discarding only incorrect content, our human verification process prioritizes selecting question-option sets that best reflect intended cultural subtleties. Although this leads to setting aside numerous plausible samples, it does not imply unreliability; instead, it refines the dataset to ensure deep cultural resonance. In particular, for \textsc{Type 2} questions, we removed those that primarily tested simple visual recognition rather than cultural knowledge application. Approved samples require minimal revision before inclusion, emphasizing nuanced cultural alignment and highlighting the need for further work to better synchronize VLM outputs with human cultural intentions.

\subsection{\texttt{K-Viscuit}}
\label{subsec:data_analysis}

\noindent\textbf{{Statistics}}
Table~\ref{tab:data_stat} presents detailed statistics of our benchmark dataset.
Our dataset comprises 657 total examples (237 \textsc{Type 1} and 420 \textsc{Type 2} questions) based on 237 unique images across 10 concept categories.
The average word counts are 10.11 and 15.46 for \textsc{Type 1} and \textsc{Type 2} questions respectively, with an overall average of 13.53 words.
Each question includes four options, totaling 2,628 options with an average length of 1.74 words.
Fig.~\ref{fig:pie_chart} shows the distribution of concept categories in our dataset.
We also compare our dataset size with CVQA~\citep{romero2024cvqa}, a recently proposed cultural VQA benchmark, in Appendix~\ref{sec:cvqa_comparison}.

\noindent\textbf{{Required Knowledge to Solve Questions}}
To characterize our dataset, we analyze the types of knowledge required to solve questions.
Following \citet{tong2024eyes}, we used \texttt{GPT-4} to analyze \textsc{Type 2} questions.
We provided all \textsc{Type 2} QA pairs and summarized the required knowledge for each question.
As shown in Fig.~\ref{fig:cultural_knowledge}, the analysis reveals diverse cultural elements.
Detailed categorization instructions are provided in Appendix~\ref{sec:app_require_knowledge_detail}.

\noindent\textbf{{Qualitative Examples}}
Fig.~\ref{fig:data_example} showcases sample images, questions, and options along with their concept categories and question types.
\section{Experiments}
\label{sec:experiment}
This section presents our experiments, including VLM evaluation, human benchmarking, and the impact of question language on accuracy.

\subsection{Models}
The following open-source VLMs are used for experiments: InstructBLIP-7B/13B~\citep{dai2024instructblip}, LLaVA-v1.6-7B/13B~\citep{liu2024visual}, mPLUG-Owl2-7B~\citep{ye2023mplug}, InternLM-XComposer2-VL-7B~\citep{dong2024internlm}, Molmo-7B-D \citep{deitke2024molmo}, Idefics2-8B~\cite{2024Idefics2}, and Llama-3.2-11B-Vision-Instruct \citep{dubey2024llama}. 
We also use the following proprietary models: Claude-3-opus~\citep{anthropic2024claude}, GPT-4-Turbo~\citep{gpt4}, Gemini-1.5-Pro~\citep{reid2024gemini}, and GPT-4o \citep{openai2024gpt4o}. 
All models are evaluated in the conventional multiple-choice setup, where a model is prompted to choose its answer from one of four options.
The input text is constructed by concatenating (1) a question, (2) each option with option letters in alphabetical order, and (3) the instruction about output format (i.e., \textit{"Answer with the option's letter from the given choices directly."}).
We use accuracy as an evaluation metric.

\subsection{Results and Analyses}
\paragraph{Main Results} 
Table \ref{tab:main_table} demonstrates the evaluation results of different VLMs on our dataset. 
We first observe that proprietary models usually show higher accuracies. 
The GPT-4o and Gemini-1.5-Pro achieve the highest and second-highest scores, respectively. Among the open-sourced models, Llama-3.2-11B and Idefics2-8B usually perform better than other models.
The accuracy of models in different question types is shown in Table~\ref{tab:type_table}.
We observe that most models show higher accuracy in \textsc{Type 2} questions compared to \textsc{Type 1} questions.
Regarding these trends, we suspect visual recognition with diverse cultural contexts poses inherent challenges for VLMs.\footnote{Further analysis of this trend is provided in \ref{sec:app:type1-shot}.} 
Our dataset enables evaluating and identifying such aspects where models can be improved.

\begin{table}[t!]
\small
\centering

\begin{tabular}{l|cc|c}
\toprule

\textbf{Model} & \textbf{\textsc{Type 1}} & \textbf{\textsc{Type 2}} & \textbf{All} \\
\midrule
InstructBLIP-7B & 45.57 & 53.81 & 50.84 \\
InstructBLIP-13B & 51.48 & 57.86 & 55.56 \\
mPLUG-Owl2-7B & 43.04 & 51.19  & 48.25 \\
LLaVA-1.6-7B & 50.21 & 59.76 & 56.32 \\
LLaVA-1.6-13B & 54.01 & 58.81 & 57.08 \\
InternLM-XC2-7B & 56.12 & 61.67 & 59.67 \\
Molmo-7B-D&     55.27 & 64.29	&61.04 \\
Idefics2-8B &   63.71 & 63.57 & 63.62 \\
Llama-3.2-11B & 69.20 &	67.38&	68.04 \\
\midrule
Claude-3-opus & 69.62 & 80.24 & 70.02               \\
GPT-4-Turbo & 78.90 & \underline{81.90} & 80.82 \\
Gemini-1.5-Pro & \underline{83.97} & 70.24 & \underline{81.58} \\
GPT-4o & \textbf{92.41}	& \textbf{87.86} &	\textbf{89.50} \\
\bottomrule
\end{tabular}
\caption{\textbf{Results on Different Question Types.} \textsc{Type 1} and \textsc{Type 2} denote question types in visual recognition and cultural knowledge application, respectively.}
\label{tab:type_table}

\end{table}

\begin{figure}[t!]
\centering
\includegraphics[width=0.85\columnwidth]{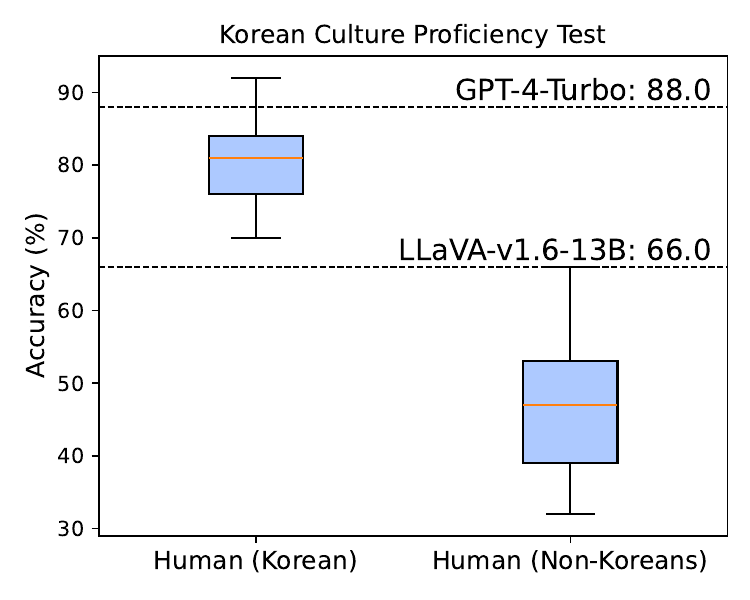}
\vspace{-0.3cm}
\caption{\textbf{Human Evaluation.} Accuracy comparison between Koreans and non-Koreans on 50 samples of \texttt{K-Viscuit}. The performance of selected models on this subset is also displayed. The average scores were 80.2 (Korean participants) and 47.0 (non-Korean participants), with standard deviations of 2.69 and 5.95.}
\label{fig:user_study_figure}
\vspace{-0.4cm}
\end{figure}

\paragraph{Human Evaluation on the Benchmark}
We conducted a human evaluation to evaluate how well people from different backgrounds perform on \texttt{K-Viscuit}.
We selected a subset of our dataset by randomly sampling 25 images, each with one \textsc{type 1} and one \textsc{type 2} question, totaling 50 questions.
This test was administered to 20 Koreans and 14 non-Koreans.
As depicted in Fig.~\ref{fig:user_study_figure}, the results showed that participants with better knowledge of Korean culture achieved higher accuracy.
However, even among Koreans, knowledge gaps exist between individuals and within a single person's knowledge across different domains, such as history.
A Proprietary VLM (i.e., GPT-4-Turbo) shows comparable performance to human performance, indicating that utilizing VLMs for generating diverse, culturally nuanced questions is more effective than relying solely on individual human annotators. 
Details of the user study are provided in the Appendix~\ref{sec:app_user_study}.

\begin{table}[t!]
\small
\centering

\begin{tabular}{l|ccc}
\toprule

\textbf{Model} &  \textsc{en} & \textsc{ko}$^*$ &\textsc{en} + \textsc{ko$^*$} \\
\midrule
Claude-3-Opus   & 70.02 & 65.30 & 70.32\\
GPT-4-Turbo     & 80.82 & 75.34 & 80.37\\
Gemini-1.5-Pro  & 81.58 & 80.82 & 83.41\\
GPT-4o          & 89.50 & 76.56 & 79.76 \\

\bottomrule
\end{tabular}
\caption{\textbf{Results with Different Input Languages}. \textsc{ko}$^*$ is machine-translated texts.
For \textsc{en}+\textsc{ko$^*$}, we provide questions and options in both languages to models.}
\label{tab:transaltion}
\vspace{-0.2cm}
\end{table}
\begin{figure*}[t!]
\centering
\includegraphics[width=0.95\textwidth]{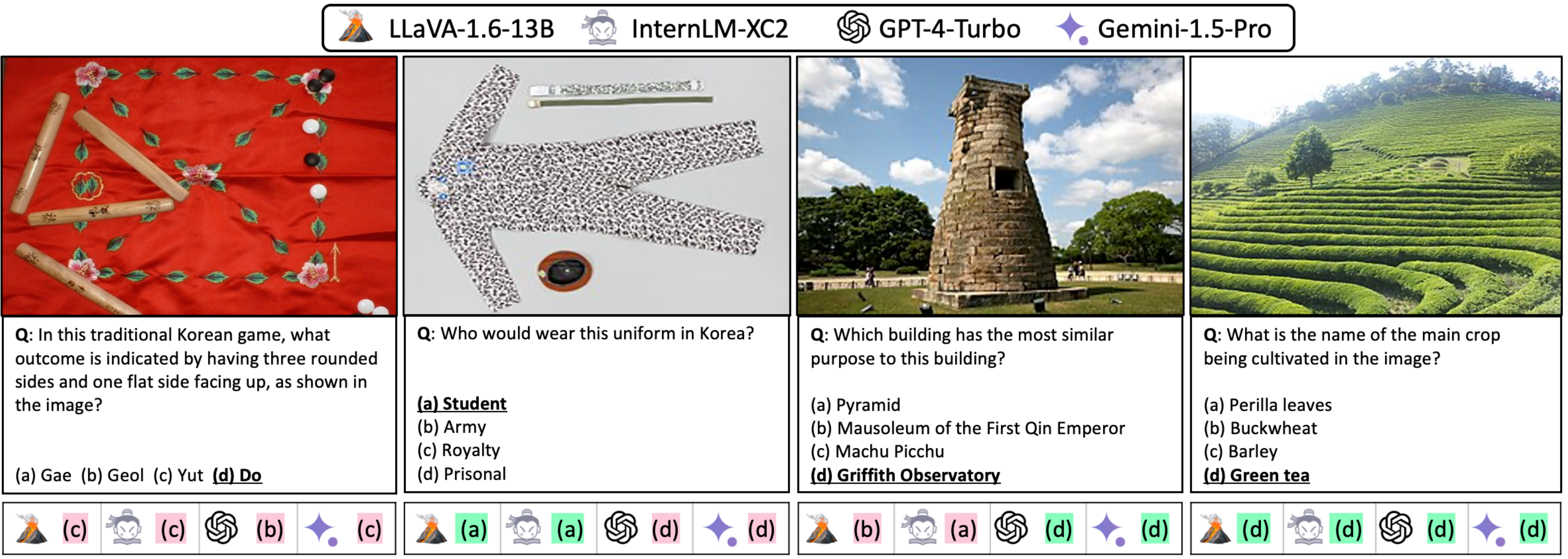}
\caption{Qualitative examples of selected VLMs (i.e., LLaVA-1.6-13B, InternLM-XC2-7B, GPT-4-Turbo, and Gemini-1.5-Pro) on sampled questions. The answer option is highlighted in the underline. The correct and incorrect choice of models are highlighted in \colorbox{green!30}{green} and \colorbox{pink!30}{red}.}
\label{fig:prediction_sample_fig}
\vspace{-0.3cm}
\end{figure*}

\paragraph{Asking the VLM in Korean} 
In this work, we primarily focused on evaluating the VLM's understanding of Korean culture and intentionally excluded their understanding of the Korean language.
Thus, we designed the dataset to focus on the VLM's multiculturality, independent of its multilingualism.
However, cultural information about certain groups often exists in the language that persons in the group frequently use.
In other words, VLMs trained in multilingual corpora might have learned about Korean culture through texts written in Korean.
Therefore, we probe whether asking questions in Korean could improve the performance of VLMs on our dataset.
To translate our dataset into Korean, we use proprietary unimodal LM (\texttt{gpt-3.5-turbo}) as a machine translation system.
For each sample in our dataset, the LM translates questions and four options into Korean.
Three proprietary VLMs that can receive Korean text (i.e., GPT-4-Turbo, Claude-3-opus, and Gemini-1.5-Pro) are used for experiments. 

Evaluation results with different input languages are presented in Table~\ref{tab:transaltion}.
We observe that solely providing translated texts in Korean to VLMs does not contribute to model performance.
When English texts are also given to models, Gemini-1.5-Pro shows increased performance (i.e., 81.58 to 83.41).
GPT-4-Turbo and Claude-3-opus usually do not take the benefits of using Korean texts.

\begin{table}[t]
\small
\centering
\begin{tabular}{lccc}
\toprule
 & \multicolumn{1}{l}{\textbf{VQAv2}} & \multicolumn{1}{l}{\textbf{CVQA}} & \multicolumn{1}{l}{\textbf{K-Viscuit}} \\ \midrule
LLaVA-v1.6-13B & 82.8 & 57.93 & 57.08 \\
Molmo-7B-D     & 85.6 & 65.52 & 61.04 \\
Idefics2-8b    & 81.2 & 68.97 & 63.62 \\
Llama-3.2-11B  & 75.2 & 72.41 & 68.04 \\ \bottomrule
\end{tabular}
\vspace{-0.1cm}
\caption{Evaluation results of selected open-source VLMs on various VQA datasets. The results on the VQAv2 \citep{goyal2017making} are taken from the respective papers of each model. The performance on the CVQA dataset was measured on the Korean subset.}
\label{tab:compare_w_other}
\vspace{-0.4cm}
\end{table}

\paragraph{Comparison of open-source VLMs on Various VQA Datasets}
We conducted experiments to measure the performance of VLMs on various VQA datasets. To this end, we compared the performance of four open-source VLMs on commonly used VQA benchmarks: the VQAv2 \citep{goyal2017making} dev-test split, the Korean subset of CVQA \citep{romero2024cvqa}, and our dataset, with results shown in Table \ref{tab:compare_w_other}. The VLMs demonstrated their best performance on VQAv2, while showing relatively lower accuracy on CVQA and our dataset. This suggests that the cultural questions we collected are indeed challenging for VLMs to solve.

\begin{figure*}[t!]
\centering
\includegraphics[width=2\columnwidth]{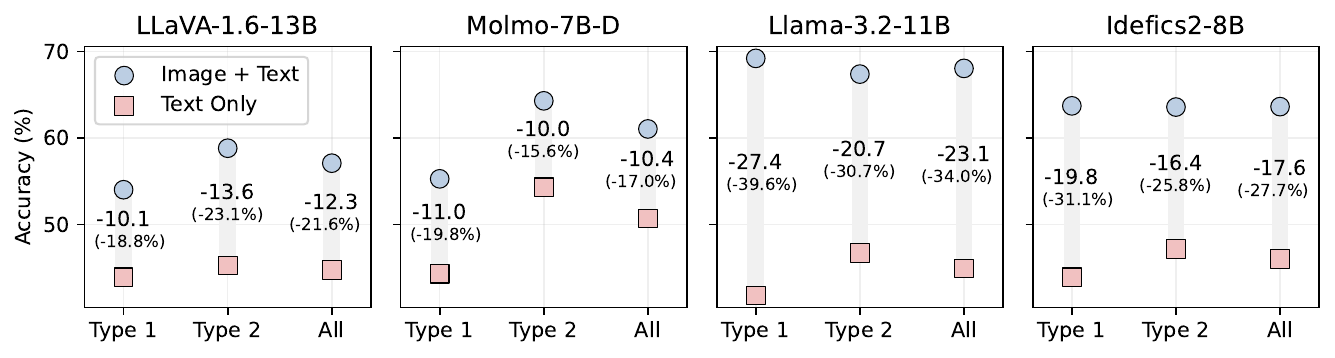}
\vspace{-0.1cm}
\caption{\textbf{Visual Dependency Analysis.} Accuracy comparison of selected models when using actual images versus noise images. The numbers indicate absolute and relative accuracy drops when replacing images with noise.}
\vspace{-0.2cm}
\label{fig:white_image}
\end{figure*}

\paragraph{Evaluating the Visual Dependency of K-Viscuit Questions}

To assess how much the K-Viscuit dataset relies on visual information, we conducted an experiment measuring the impact of removing meaningful image content. Specifically, models are provided with randomly generated images with pixel values sampled from a Gaussian distribution instead of actual images. These images are of the same size as the originals but contain no meaningful visual structure. By comparing model performance in this setting to standard image-based evaluation, we can determine the extent to which K-Viscuit questions require visual understanding rather than relying solely on textual cues.

Fig. \ref{fig:white_image} presents the accuracy of selected models under two conditions: with \textit{actual images (Image + Text)} and with \textit{noise images (Text Only)}. Across all models, a substantial drop in accuracy is observed when images are replaced with noise, confirming that K-Viscuit questions require meaningful visual information. The performance degradation varies by model, with Llama-3.2-11B showing the largest accuracy drop, while Molmo-7B-D is comparatively less affected. Additionally, both Type 1 and Type 2 questions exhibit large performance declines in the absence of images, suggesting that visual information is crucial across different question types. These results highlight the varying degrees of visual dependence across models and reinforce the necessity of strong visual reasoning capabilities for solving K-Viscuit questions.

\paragraph{Qualitative Results}
Fig.~\ref{fig:prediction_sample_fig} presents the prediction of selected models on sampled examples.
In the first example, all models fail to correctly answer the question about asking detailed game rules.
In the second case, two proprietary models are wrong while open-source models make correct answers.
The third problem requires the model to recognize the structure in the image as Cheomseongdae and infer that it serves a function similar to that of the Griffith Observatory in the United States.
Both proprietary models make correct answers to this example.
In the final example, all models successfully identify the correct answer about agriculture.

\section{Future Directions}
This section explores future directions for improving model evaluation (\S\ref{sec:mcqa-gqa}) and enhancing model performance with external knowledge (\S\ref{sec:rag}).

\subsection{Evaluation beyond Multiple-choice VQA}
\label{sec:mcqa-gqa}
\begin{figure}[t!]
\centering
\includegraphics[width=\columnwidth]{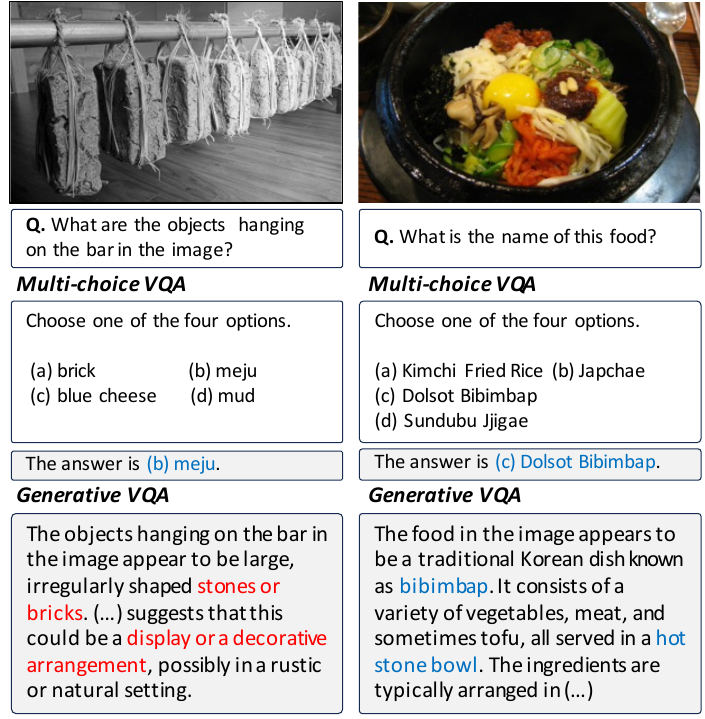}
\vspace{-0.5cm}
\caption{Qualitative results of a VLM with multi-choice and generative VQA setups. LLaVA-1.6-13B is used for the analysis. The \textcolor{blue}{correct} and \textcolor{red}{incorrect} model generations are manually highlighted by the authors.}
\vspace{-0.45cm}
\label{fig:generative_example}
\end{figure}

Our dataset is designed as a multiple-choice Visual Question Answering (VQA) task, where Vision-Language Models (VLMs) select one answer from given options. While this classification setup enables straightforward performance measurement through accuracy, it may not fully reveal the models' depth of cultural understanding. For instance, VLMs may correctly select an answer from the available options but fail in a generative VQA setup, where they must generate a free-form answer. As shown in Fig.~\ref{fig:generative_example}, while the VLM accurately identifies the object as \textit{bibimbap} in both multiple-choice and generative setups for the first example, it fails to provide accurate cultural context in the second example despite choosing the correct option in the multiple-choice format.

\begin{table}[t]
\begin{adjustbox}{width=\columnwidth}
\begin{tabular}{cccc}
\toprule
\multicolumn{1}{c}{\textbf{Correct}} & \multicolumn{1}{c}{\textbf{Hallucinated}} & \multicolumn{1}{c}{\textbf{Abstained}} & \multicolumn{1}{c}{\textbf{Total}} \\ \midrule
29 (36.35\%) & 34 (42.5\%) & 17 (21.25\%) & 80 (100\%) \\
\bottomrule
\end{tabular}
\end{adjustbox}
\vspace{-0.1cm}
\caption{\textbf{Generative QA Results.} We manually evaluated the responses of LLaVA-1.6-13B on \textit{Food} category.}
\label{tab:generative_table}
\vspace{-0.4cm}
\end{table}

To quantitatively analyze this aspect, we randomly sampled 80 questions from the Food category and evaluated Llava-v1.6-13B's performance in a generative setting. The model was prompted with the instruction: \textit{"This is a question about Korean culture. Answer the given questions briefly and concisely."} without access to the original multiple-choice options. Two native Korean evaluators assessed the generated responses using three categories: (1) \textit{Correct}: The model's response fully contained the original answer or was accurate to the question, (2) \textit{Hallucinated}: The model provided incorrect information, and (3) \textit{Abstained}: The model's response did not include a direct answer to the question but provided a relevant description of the image and did not attempt to generate an answer hastily (e.g., \textit{"The specific type of food and its cultural context would determine when it is commonly served"}).

The human analysis results are shown in Table~\ref{tab:generative_table}. We observed that the VLM's accuracy in the generative setup was relatively lower than in the multiple-choice setup (from 45.07\% to 36.25\%). This decrease in accuracy is likely because the task is inherently more difficult without predefined options. Interestingly, a large proportion of responses (21.25\%) were classified as abstained, indicating that the VLM recognized the need for additional information to answer confidently and refrained from providing a definitive but potentially incorrect response. This behavior suggests that providing the necessary external knowledge could enhance performance. Our future research aims to propose various evaluation methods to assess whether VLMs genuinely possess the knowledge to answer questions accurately.

\subsection{Augmenting VLMs with External Knowledge}
\label{sec:rag}

In previous experiments, models struggled with questions requiring cultural knowledge.
To address these gaps, we considered fine-tuning open-source models and augmenting models with external knowledge.
Since the latter can be applied to both open-source and proprietary models, we focused on that approach.
We augmented the models with relevant documents for each test image from the \textsc{Food} concept.
Our retrieval method involved generating captions using GPT-4-Turbo, embedding these captions via \texttt{text-embedding-3-large} to build queries, and retrieving Top-1 document from 152 Wikipedia pages related to the objects mentioned in the \textsc{Type 1} options.
The \texttt{K-Viscuit} distractors were designed to closely resemble correct answers, creating a challenging retrieval setting with numerous hard negatives that mimic real-world conditions.
We assess if the retrieval method remains effective under these conditions by examining performance changes when cultural knowledge is introduced.

As shown in Table \ref{table:retreival}, providing retrieved documents can enhance model performance in the food category, as demonstrated by improved scores in the \textbf{\textsc{Retrieved}} setting compared to \textbf{\textsc{None}}. In cases where proprietary models did not benefit from certain retrieved documents, their performance still surpassed the baseline when given carefully curated \textbf{\textsc{Oracle}} documents. This suggests that retrieval-based augmentation has the potential to strengthen cultural knowledge in VQA tasks, provided that the quality and relevance of the selected documents are refined.
We provide the details in the Appendix~\ref{sec:app_retrieval}.

\begin{table}[t!]
\centering
\small
\begin{adjustbox}{width=\columnwidth}
\begin{tabular}{l|cc|c}
\toprule
\textbf{Model} & \textbf{\textsc{None}} & \textbf{\textsc{Retrieved}} & \textbf{\textsc{Oracle}}  \\
\midrule
LLaVA-1.6-7B & 43.66 & 68.31 & 78.87 \\
LLaVA-1.6-13B & 45.77 & 64.08 & 80.28 \\
InternLM-XC2-7B & 50.70 & 68.31 & 82.39 \\
Idefics2-8B & 51.41 & 67.61 & 82.39 \\
\midrule
Claude-3-opus & 62.68 & 70.42 & 87.32 \\
GPT-4-Turbo & 73.94 & 78.17 & 88.73 \\
Gemini-1.5-Pro & 80.28 & 78.17 & 90.85 \\
GPT-4o & 88.73 & 83.10 & 92.25 \\
\bottomrule
\end{tabular}
\end{adjustbox}
\caption{Retrieval-augmented generation results on \textit{Food} category. Oracle documents were manually selected by the authors.}
\label{table:retreival}
\vspace{-0.5cm}
\end{table}

\section{Conclusion}
\label{sec:conclusion}

In this work, we proposed a semi-automated framework for constructing culturally aware benchmarks for vision-language models through human-VLM collaboration, focusing on visual recognition and cultural knowledge application in multi-choice QA. We demonstrated its effectiveness by creating \texttt{K-Viscuit}, revealing a performance gap between proprietary and open-source VLMs in identifying Korean cultural concepts. Through detailed analyses and knowledge augmentation experiments, we specified the impact of cultural understanding on VQA performance and extended our investigation to open-ended answer generation, discussing its limitations. This work highlights the importance of cultural diversity in model evaluation and paves the way for more inclusive VLMs.

\section*{Limitations}
\label{sec:limitations}
Our current framework requires a manual selection of images that match specified concepts, preventing fully automated dataset generation. 
These human efforts can be alleviated by multimodal retrieval modules to some extent. 
To this end, it should be predetermined whether current multimodal encoders can sufficiently understand culturally nuanced images and texts. 
Enhancing retrieval models to better understand and match cultural contexts remains our exciting future work.
The manual verification of automatically generated questions also can be a considerable burden. Developing a quality estimation module for generated questions could assist in this process by reducing the workload on human annotators.
Additionally, LLMs are highly sensitive to the order of answer options in multiple-choice QA tasks, as discussed in \citet{positionbias,pezeshkpour-hruschka-2024-large}. While we mitigate positional bias in model evaluation by shuffling the answer options, this approach may not be entirely sufficient. As shown in \ref{app:sec:positional}, VLMs often fail to produce consistent predictions when the positions of the answer options are altered. Future work could explore calibration techniques, such as contrastive decoding or uncertainty-aware selection strategies, to further enhance robustness in multiple-choice settings.

\section*{Ethical Statement}
\label{sec:ethical}
In constructing K-Viscuit, we ensured all images were sourced from Wikimedia Commons under Creative Commons licenses, maintaining proper attribution and copyright compliance. The dataset was carefully curated to avoid harmful or inappropriate content, and all questions and answers were verified by native Korean speakers to ensure cultural relevance. While comprehensive, we acknowledge that our dataset captures only a subset of Korean cultural elements.

\section*{Acknowledgements}
This work was supported by a grant of the KAIST-KT joint research project through AI2X Laboratory, Tech Innovation Group , funded by KT (No. D23000019, Research on Multilingual Multicultural Vision-Language Representation Learning) and Institute for Information \& communications Technology Planning \& Evaluation(IITP) grant funded by the Korea government(MSIT) (RS-2019-II190075, Artificial Intelligence Graduate School Program(KAIST) and RS-2025-02304967, AI Star Fellowship(KAIST).


\bibliography{custom}

\begin{thebibliography}{40}
\providecommand{\natexlab}[1]{#1}

\bibitem[{Achiam et~al.(2023)Achiam, Adler, Agarwal, Ahmad, Akkaya, Aleman, Almeida, Altenschmidt, Altman, Anadkat et~al.}]{gpt4}
Josh Achiam, Steven Adler, Sandhini Agarwal, Lama Ahmad, Ilge Akkaya, Florencia~Leoni Aleman, Diogo Almeida, Janko Altenschmidt, Sam Altman, Shyamal Anadkat, et~al. 2023.
\newblock Gpt-4 technical report.
\newblock \emph{arXiv preprint arXiv:2303.08774}.

\bibitem[{Anthropic(2024)}]{anthropic2024claude}
AI~Anthropic. 2024.
\newblock The claude 3 model family: Opus, sonnet, haiku.
\newblock \emph{Claude-3 Model Card}.

\bibitem[{Antol et~al.(2015)Antol, Agrawal, Lu, Mitchell, Batra, Zitnick, and Parikh}]{antol2015vqa}
Stanislaw Antol, Aishwarya Agrawal, Jiasen Lu, Margaret Mitchell, Dhruv Batra, C~Lawrence Zitnick, and Devi Parikh. 2015.
\newblock Vqa: Visual question answering.
\newblock In \emph{Proceedings of the IEEE international conference on computer vision}, pages 2425--2433.

\bibitem[{Bartolo et~al.(2022)Bartolo, Thrush, Riedel, Stenetorp, Jia, and Kiela}]{bartolo2022models}
Max Bartolo, Tristan Thrush, Sebastian Riedel, Pontus Stenetorp, Robin Jia, and Douwe Kiela. 2022.
\newblock Models in the loop: Aiding crowdworkers with generative annotation assistants.
\newblock In \emph{Proceedings of the 2022 Conference of the North American Chapter of the Association for Computational Linguistics: Human Language Technologies}, pages 3754--3767.

\bibitem[{Bhatia et~al.(2024)Bhatia, Ravi, Chinchure, Hwang, and Shwartz}]{bhatia-etal-2024-local}
Mehar Bhatia, Sahithya Ravi, Aditya Chinchure, EunJeong Hwang, and Vered Shwartz. 2024.
\newblock \href {https://doi.org/10.18653/v1/2024.emnlp-main.385} {From local concepts to universals: Evaluating the multicultural understanding of vision-language models}.
\newblock In \emph{Proceedings of the 2024 Conference on Empirical Methods in Natural Language Processing}, pages 6763--6782, Miami, Florida, USA. Association for Computational Linguistics.

\bibitem[{Dai et~al.(2024)Dai, Li, Li, Tiong, Zhao, Wang, Li, Fung, and Hoi}]{dai2024instructblip}
Wenliang Dai, Junnan Li, Dongxu Li, Anthony Meng~Huat Tiong, Junqi Zhao, Weisheng Wang, Boyang Li, Pascale~N Fung, and Steven Hoi. 2024.
\newblock Instructblip: Towards general-purpose vision-language models with instruction tuning.
\newblock \emph{Advances in Neural Information Processing Systems}, 36.

\bibitem[{Deitke et~al.(2024)Deitke, Clark, Lee, Tripathi, Yang, Park, Salehi, Muennighoff, Lo, Soldaini et~al.}]{deitke2024molmo}
Matt Deitke, Christopher Clark, Sangho Lee, Rohun Tripathi, Yue Yang, Jae~Sung Park, Mohammadreza Salehi, Niklas Muennighoff, Kyle Lo, Luca Soldaini, et~al. 2024.
\newblock Molmo and pixmo: Open weights and open data for state-of-the-art multimodal models.
\newblock \emph{arXiv preprint arXiv:2409.17146}.

\bibitem[{Dong et~al.(2024)Dong, Zhang, Zang, Cao, Wang, Ouyang, Wei, Zhang, Duan, Cao et~al.}]{dong2024internlm}
Xiaoyi Dong, Pan Zhang, Yuhang Zang, Yuhang Cao, Bin Wang, Linke Ouyang, Xilin Wei, Songyang Zhang, Haodong Duan, Maosong Cao, et~al. 2024.
\newblock Internlm-xcomposer2: Mastering free-form text-image composition and comprehension in vision-language large model.
\newblock \emph{arXiv preprint arXiv:2401.16420}.

\bibitem[{Dubey et~al.(2024)Dubey, Jauhri, Pandey, Kadian, Al-Dahle, Letman, Mathur, Schelten, Yang, Fan et~al.}]{dubey2024llama}
Abhimanyu Dubey, Abhinav Jauhri, Abhinav Pandey, Abhishek Kadian, Ahmad Al-Dahle, Aiesha Letman, Akhil Mathur, Alan Schelten, Amy Yang, Angela Fan, et~al. 2024.
\newblock The llama 3 herd of models.
\newblock \emph{arXiv preprint arXiv:2407.21783}.

\bibitem[{Dubois et~al.(2024)Dubois, Li, Taori, Zhang, Gulrajani, Ba, Guestrin, Liang, and Hashimoto}]{dubois2024alpacafarm}
Yann Dubois, Chen~Xuechen Li, Rohan Taori, Tianyi Zhang, Ishaan Gulrajani, Jimmy Ba, Carlos Guestrin, Percy~S Liang, and Tatsunori~B Hashimoto. 2024.
\newblock Alpacafarm: A simulation framework for methods that learn from human feedback.
\newblock \emph{Advances in Neural Information Processing Systems}, 36.

\bibitem[{Goyal et~al.(2017)Goyal, Khot, Summers-Stay, Batra, and Parikh}]{goyal2017making}
Yash Goyal, Tejas Khot, Douglas Summers-Stay, Dhruv Batra, and Devi Parikh. 2017.
\newblock Making the v in vqa matter: Elevating the role of image understanding in visual question answering.
\newblock In \emph{Proceedings of the IEEE conference on computer vision and pattern recognition}, pages 6904--6913.

\bibitem[{Kamalloo et~al.(2023)Kamalloo, Jafari, Zhang, Thakur, and Lin}]{kamalloo2023hagrid}
Ehsan Kamalloo, Aref Jafari, Xinyu Zhang, Nandan Thakur, and Jimmy Lin. 2023.
\newblock Hagrid: A human-llm collaborative dataset for generative information-seeking with attribution.
\newblock \emph{arXiv preprint arXiv:2307.16883}.

\bibitem[{Key and Comrie(2015)}]{key_comrie_2015}
Mary~Ritchie Key and Bernard Comrie, editors. 2015.
\newblock \emph{{IDS}}.
\newblock Max Planck Institute for Evolutionary Anthropology, Leipzig.

\bibitem[{Kim et~al.(2024)Kim, Suk, Oh, Yoo, Thorne, and Oh}]{kim2024click}
Eunsu Kim, Juyoung Suk, Philhoon Oh, Haneul Yoo, James Thorne, and Alice Oh. 2024.
\newblock Click: A benchmark dataset of cultural and linguistic intelligence in korean.
\newblock In \emph{Proceedings of the 2024 Joint International Conference on Computational Linguistics, Language Resources and Evaluation (LREC-COLING 2024)}, pages 3335--3346.

\bibitem[{Kim et~al.(2023)Kim, Bae, Shin, Kang, Kwak, Yoo, and Seo}]{kim2023aligning}
Sungdong Kim, Sanghwan Bae, Jamin Shin, Soyoung Kang, Donghyun Kwak, Kang Yoo, and Minjoon Seo. 2023.
\newblock Aligning large language models through synthetic feedback.
\newblock In \emph{Proceedings of the 2023 Conference on Empirical Methods in Natural Language Processing}, pages 13677--13700.

\bibitem[{Kramsch(2014)}]{kramsch2014language}
Claire Kramsch. 2014.
\newblock Language and culture.
\newblock \emph{AILA review}, 27(1):30--55.

\bibitem[{Lauren{\c{c}}on et~al.(2024)Lauren{\c{c}}on, Tronchon, Cord, and Sanh}]{2024Idefics2}
Hugo Lauren{\c{c}}on, L{\'e}o Tronchon, Matthieu Cord, and Victor Sanh. 2024.
\newblock What matters when building vision-language models?
\newblock \emph{arXiv preprint arXiv:2405.02246}.

\bibitem[{Lin et~al.(2014)Lin, Maire, Belongie, Hays, Perona, Ramanan, Doll{\'a}r, and Zitnick}]{lin2014microsoft}
Tsung-Yi Lin, Michael Maire, Serge Belongie, James Hays, Pietro Perona, Deva Ramanan, Piotr Doll{\'a}r, and C~Lawrence Zitnick. 2014.
\newblock Microsoft coco: Common objects in context.
\newblock In \emph{Computer Vision--ECCV 2014: 13th European Conference, Zurich, Switzerland, September 6-12, 2014, Proceedings, Part V 13}, pages 740--755. Springer.

\bibitem[{Liu et~al.(2022)Liu, Swayamdipta, Smith, and Choi}]{liu2022wanli}
Alisa Liu, Swabha Swayamdipta, Noah~A Smith, and Yejin Choi. 2022.
\newblock Wanli: Worker and ai collaboration for natural language inference dataset creation.
\newblock In \emph{Findings of the Association for Computational Linguistics: EMNLP 2022}, pages 6826--6847.

\bibitem[{Liu et~al.(2021)Liu, Bugliarello, Ponti, Reddy, Collier, and Elliott}]{marvl}
Fangyu Liu, Emanuele Bugliarello, Edoardo~Maria Ponti, Siva Reddy, Nigel Collier, and Desmond Elliott. 2021.
\newblock Visually grounded reasoning across languages and cultures.
\newblock In \emph{Proceedings of the 2021 Conference on Empirical Methods in Natural Language Processing}, pages 10467--10485.

\bibitem[{Liu et~al.(2024{\natexlab{a}})Liu, Li, Wu, and Lee}]{liu2024visual}
Haotian Liu, Chunyuan Li, Qingyang Wu, and Yong~Jae Lee. 2024{\natexlab{a}}.
\newblock llava.
\newblock \emph{Advances in neural information processing systems}, 36.

\bibitem[{Liu et~al.(2024{\natexlab{b}})Liu, Duan, Zhang, Li, Zhang, Zhao, Yuan, Wang, He, Liu et~al.}]{liu2024mmbench}
Yuan Liu, Haodong Duan, Yuanhan Zhang, Bo~Li, Songyang Zhang, Wangbo Zhao, Yike Yuan, Jiaqi Wang, Conghui He, Ziwei Liu, et~al. 2024{\natexlab{b}}.
\newblock Mmbench: Is your multi-modal model an all-around player?
\newblock In \emph{European conference on computer vision}, pages 216--233. Springer.

\bibitem[{Nayak et~al.(2024)Nayak, Jain, Awal, Reddy, Steenkiste, Hendricks, Stanczak, and Agrawal}]{nayak2024benchmarking}
Shravan Nayak, Kanishk Jain, Rabiul Awal, Siva Reddy, Sjoerd Steenkiste, Lisa Hendricks, Karolina Stanczak, and Aishwarya Agrawal. 2024.
\newblock Benchmarking vision language models for cultural understanding.
\newblock In \emph{Proceedings of the 2024 Conference on Empirical Methods in Natural Language Processing}, pages 5769--5790.

\bibitem[{OpenAI(2024)}]{openai2024gpt4o}
OpenAI. 2024.
\newblock \href {https://openai.com/index/hello-gpt-4o/} {Hello gpt-4o: Enhanced efficiency for gpt models}.
\newblock Accessed: 2024-05-13.

\bibitem[{Pezeshkpour and Hruschka(2024)}]{pezeshkpour-hruschka-2024-large}
Pouya Pezeshkpour and Estevam Hruschka. 2024.
\newblock \href {https://doi.org/10.18653/v1/2024.findings-naacl.130} {Large language models sensitivity to the order of options in multiple-choice questions}.
\newblock In \emph{Findings of the Association for Computational Linguistics: NAACL 2024}, pages 2006--2017, Mexico City, Mexico. Association for Computational Linguistics.

\bibitem[{Ramaswamy et~al.(2024)Ramaswamy, Lin, Zhao, Adcock, van~der Maaten, Ghadiyaram, and Russakovsky}]{ramaswamy2024geode}
Vikram~V Ramaswamy, Sing~Yu Lin, Dora Zhao, Aaron Adcock, Laurens van~der Maaten, Deepti Ghadiyaram, and Olga Russakovsky. 2024.
\newblock Geode: a geographically diverse evaluation dataset for object recognition.
\newblock \emph{Advances in Neural Information Processing Systems}, 36.

\bibitem[{Ramos(2020)}]{ramos2020cognitive}
Hector Ramos. 2020.
\newblock Cognitive fixation and creativity.
\newblock In \emph{Encyclopedia of creativity, invention, innovation and entrepreneurship}, pages 319--320. Springer.

\bibitem[{Reid et~al.(2024)Reid, Savinov, Teplyashin, Lepikhin, Lillicrap, Alayrac, Soricut, Lazaridou, Firat, Schrittwieser et~al.}]{reid2024gemini}
Machel Reid, Nikolay Savinov, Denis Teplyashin, Dmitry Lepikhin, Timothy Lillicrap, Jean-baptiste Alayrac, Radu Soricut, Angeliki Lazaridou, Orhan Firat, Julian Schrittwieser, et~al. 2024.
\newblock Gemini 1.5: Unlocking multimodal understanding across millions of tokens of context.
\newblock \emph{arXiv preprint arXiv:2403.05530}.

\bibitem[{Romero et~al.(2024)Romero, Lyu, Wibowo, Lynn, Hamed, Kishore, Mandal, Dragonetti, Abzaliev, Tonja, Balcha, Whitehouse, Salamea, Velasco, Adelani, Meur, Villa-Cueva, Koto, Farooqui, Belcavello, Batnasan, Vallejo, Caulfield, Ivetta, Song, Ademtew, Maina, Lovenia, Azime, Cruz, Gala, Geng, Ortiz-Barajas, Baek, Dunstan, Alemany, Nagasinghe, Benotti, D'Haro, Viridiano, Estecha-Garitagoitia, Cabrera, Rodríguez-Cantelar, Jouitteau, Mihaylov, Imam, Adilazuarda, Gochoo, Otgonbold, Etori, Niyomugisha, Silva, Chitale, Dabre, Chevi, Zhang, Diandaru, Cahyawijaya, Góngora, Jeong, Purkayastha, Kuribayashi, Jayakumar, Torrent, Ehsan, Araujo, Kementchedjhieva, Burzo, Lim, Yong, Ignat, Nwatu, Mihalcea, Solorio, and Aji}]{romero2024cvqa}
David Romero, Chenyang Lyu, Haryo~Akbarianto Wibowo, Teresa Lynn, Injy Hamed, Aditya~Nanda Kishore, Aishik Mandal, Alina Dragonetti, Artem Abzaliev, Atnafu~Lambebo Tonja, Bontu~Fufa Balcha, Chenxi Whitehouse, Christian Salamea, Dan~John Velasco, David~Ifeoluwa Adelani, David~Le Meur, Emilio Villa-Cueva, Fajri Koto, Fauzan Farooqui, Frederico Belcavello, Ganzorig Batnasan, Gisela Vallejo, Grainne Caulfield, Guido Ivetta, Haiyue Song, Henok~Biadglign Ademtew, Hernán Maina, Holy Lovenia, Israel~Abebe Azime, Jan Christian~Blaise Cruz, Jay Gala, Jiahui Geng, Jesus-German Ortiz-Barajas, Jinheon Baek, Jocelyn Dunstan, Laura~Alonso Alemany, Kumaranage Ravindu~Yasas Nagasinghe, Luciana Benotti, Luis~Fernando D'Haro, Marcelo Viridiano, Marcos Estecha-Garitagoitia, Maria Camila~Buitrago Cabrera, Mario Rodríguez-Cantelar, Mélanie Jouitteau, Mihail Mihaylov, Mohamed Fazli~Mohamed Imam, Muhammad~Farid Adilazuarda, Munkhjargal Gochoo, Munkh-Erdene Otgonbold, Naome Etori, Olivier Niyomugisha, Paula~Mónica Silva, Pranjal
  Chitale, Raj Dabre, Rendi Chevi, Ruochen Zhang, Ryandito Diandaru, Samuel Cahyawijaya, Santiago Góngora, Soyeong Jeong, Sukannya Purkayastha, Tatsuki Kuribayashi, Thanmay Jayakumar, Tiago~Timponi Torrent, Toqeer Ehsan, Vladimir Araujo, Yova Kementchedjhieva, Zara Burzo, Zheng~Wei Lim, Zheng~Xin Yong, Oana Ignat, Joan Nwatu, Rada Mihalcea, Thamar Solorio, and Alham~Fikri Aji. 2024.
\newblock \href {https://arxiv.org/abs/2406.05967} {Cvqa: Culturally-diverse multilingual visual question answering benchmark}.
\newblock \emph{Preprint}, arXiv:2406.05967.

\bibitem[{SUSAN(1996)}]{susan1996language}
D~SUSAN. 1996.
\newblock Language shock: understanding the culture of conversation.

\bibitem[{Taori et~al.(2023)Taori, Gulrajani, Zhang, Dubois, Li, Guestrin, Liang, and Hashimoto}]{taori2023alpaca}
Rohan Taori, Ishaan Gulrajani, Tianyi Zhang, Yann Dubois, Xuechen Li, Carlos Guestrin, Percy Liang, and Tatsunori~B Hashimoto. 2023.
\newblock Alpaca: A strong, replicable instruction-following model.
\newblock \emph{Stanford Center for Research on Foundation Models. https://crfm. stanford. edu/2023/03/13/alpaca. html}, 3(6):7.

\bibitem[{Tong et~al.(2024)Tong, Liu, Zhai, Ma, LeCun, and Xie}]{tong2024eyes}
Shengbang Tong, Zhuang Liu, Yuexiang Zhai, Yi~Ma, Yann LeCun, and Saining Xie. 2024.
\newblock Eyes wide shut? exploring the visual shortcomings of multimodal llms.
\newblock \emph{arXiv preprint arXiv:2401.06209}.

\bibitem[{Wang et~al.(2023)Wang, Chen, You, Xu, He, Li, Codella, Chang, and Chang}]{wang2023mcq}
Zhecan Wang, Long Chen, Haoxuan You, Keyang Xu, Yicheng He, Wenhao Li, Noel Codella, Kai-Wei Chang, and Shih-Fu Chang. 2023.
\newblock Dataset bias mitigation in multiple-choice visual question answering and beyond.
\newblock In \emph{Findings of the Association for Computational Linguistics: EMNLP 2023}, pages 8598--8617.

\bibitem[{Wibowo et~al.(2023)Wibowo, Fuadi, Nityasya, Prasojo, and Aji}]{wibowo2023copalID}
Haryo~Akbarianto Wibowo, Erland~Hilman Fuadi, Made~Nindyatama Nityasya, Radityo~Eko Prasojo, and Alham~Fikri Aji. 2023.
\newblock Copal-id: Indonesian language reasoning with local culture and nuances.
\newblock \emph{arXiv preprint arXiv:2311.01012}.

\bibitem[{Wolf et~al.(2020)Wolf, Debut, Sanh, Chaumond, Delangue, Moi, Cistac, Rault, Louf, Funtowicz, Davison, Shleifer, von Platen, Ma, Jernite, Plu, Xu, Scao, Gugger, Drame, Lhoest, and Rush}]{wolf-etal-2020-transformers}
Thomas Wolf, Lysandre Debut, Victor Sanh, Julien Chaumond, Clement Delangue, Anthony Moi, Pierric Cistac, Tim Rault, Rémi Louf, Morgan Funtowicz, Joe Davison, Sam Shleifer, Patrick von Platen, Clara Ma, Yacine Jernite, Julien Plu, Canwen Xu, Teven~Le Scao, Sylvain Gugger, Mariama Drame, Quentin Lhoest, and Alexander~M. Rush. 2020.
\newblock \href {https://www.aclweb.org/anthology/2020.emnlp-demos.6} {Transformers: State-of-the-art natural language processing}.
\newblock In \emph{Proceedings of the 2020 Conference on Empirical Methods in Natural Language Processing: System Demonstrations}, pages 38--45, Online. Association for Computational Linguistics.

\bibitem[{Ye et~al.(2023)Ye, Xu, Ye, Yan, Liu, Qian, Zhang, Huang, and Zhou}]{ye2023mplug}
Qinghao Ye, Haiyang Xu, Jiabo Ye, Ming Yan, Haowei Liu, Qi~Qian, Ji~Zhang, Fei Huang, and Jingren Zhou. 2023.
\newblock mplug-owl2: Revolutionizing multi-modal large language model with modality collaboration.
\newblock \emph{arXiv preprint arXiv:2311.04257}.

\bibitem[{Yin et~al.(2023)Yin, Gao, Thattai, Johnston, and Chang}]{yin2023givl}
Da~Yin, Feng Gao, Govind Thattai, Michael Johnston, and Kai-Wei Chang. 2023.
\newblock Givl: Improving geographical inclusivity of vision-language models with pre-training methods.
\newblock In \emph{Proceedings of the IEEE/CVF Conference on Computer Vision and Pattern Recognition}, pages 10951--10961.

\bibitem[{Yin et~al.(2021)Yin, Li, Hu, Peng, and Chang}]{yin2021broaden}
Da~Yin, Liunian~Harold Li, Ziniu Hu, Nanyun Peng, and Kai-Wei Chang. 2021.
\newblock Broaden the vision: Geo-diverse visual commonsense reasoning.
\newblock In \emph{Proceedings of the 2021 Conference on Empirical Methods in Natural Language Processing}, pages 2115--2129.

\bibitem[{Young et~al.(2014)Young, Lai, Hodosh, and Hockenmaier}]{young2014image}
Peter Young, Alice Lai, Micah Hodosh, and Julia Hockenmaier. 2014.
\newblock From image descriptions to visual denotations: New similarity metrics for semantic inference over event descriptions.
\newblock \emph{Transactions of the Association for Computational Linguistics}, 2:67--78.

\bibitem[{Zheng et~al.(2023)Zheng, Zhou, Meng, Zhou, and Huang}]{positionbias}
Chujie Zheng, Hao Zhou, Fandong Meng, Jie Zhou, and Minlie Huang. 2023.
\newblock Large language models are not robust multiple choice selectors.
\newblock In \emph{The Twelfth International Conference on Learning Representations}.

\end{thebibliography}
\newpage
\appendix

\section{Additional Analysis}
\label{sec:additional_analysis}

\subsection{Evaluating Sensitivity to Option Order in MCQA with Circular Evaluation}
\label{app:sec:positional}
Recent studies have shown that many vision-language models (VLMs) exhibit sensitivity to the order of answer choices in multiple-choice QA, where performance varies depending on how options are presented. To investigate this phenomenon, we apply Circular Evaluation, proposed by \citet{liu2024mmbench}. This method systematically shifts the positions of four answer choices and forwards them through the model four times, considering a prediction correct only if all variations yield the same answer.

\begin{figure}[t]
\centering
\includegraphics[width=\columnwidth]{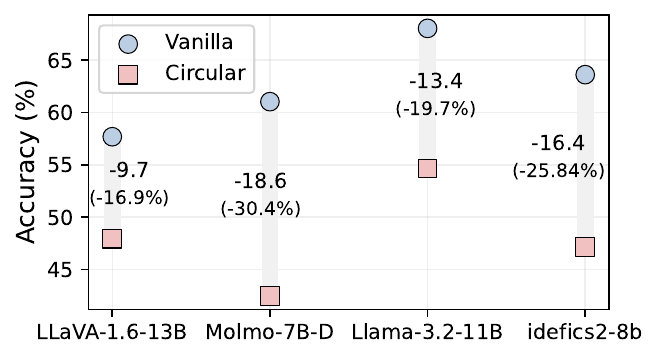}
\vspace{-0.8cm}
\caption{\textbf{Order-Sensitivity Analysis.} Accuracy comparison of selected models under vanilla and circular evaluations \citep{liu2024mmbench}. The numbers indicate the absolute and relative performance drops. }
\vspace{-0.4cm}
\label{fig:circular}
\end{figure}

Fig \ref{fig:circular} presents the accuracy of selected models under both vanilla and circular evaluations. A significant drop in performance is observed across all models when applying circular evaluation, confirming that answer order affects predictions. The relative accuracy decrease ranges from 16.9\% to 30.4\%, suggesting that models rely on positional cues rather than robust reasoning. Among the evaluated models, Molmo-7B-D exhibits the largest performance degradation, indicating heightened sensitivity to ordering effects. One potential explanation for this behavior is positional bias, where models develop a preference for certain answer positions. These results highlight sensitivity to answer order as a critical challenge in multi-choice QA for vision-language models. Future work should explore mitigation strategies, such as answer re-ranking or training objectives that explicitly reduce order sensitivity.

\subsection{Comparison of \texttt{K-Viscuit} with CVQA dataset}
\label{sec:cvqa_comparison}

Recent advances in vision-language models have sparked a growing interest in evaluating their cultural awareness across diverse global contexts. A notable contribution in this direction is CVQA~\citep{romero2024cvqa}, which introduces a comprehensive multilingual VQA benchmark spanning 28 countries with 9,044 total examples, averaging approximately 323 examples per country. Our \texttt{K-Viscuit} dataset, while focused solely on Korean culture, contains 657 examples—substantially exceeding CVQA's per-country average. Moreover, CVQA's Korean subset comprises 290 examples, less than half the size of \texttt{K-Viscuit}. This focused approach enables \texttt{K-Viscuit} to provide more comprehensive coverage of Korean cultural elements, offering deeper insights into VLMs' understanding of specific cultural contexts.

Upon reviewing the dataset split, we found that 14 answers (4.8\%) were simply "Korea," which may limit the flexibility of different prompt settings for evaluation (e.g., \textit{"This is an image taken in Korea"}), as "Korea" becomes a shortcut response. Additionally, CVQA occasionally includes distractors that are inconsistent, such as listing (a) \textit{"Gwangju,"} (b) \textit{"Seoul,"} (c) \textit{"Seattle,"} and (d) \textit{"Busan,"} where (c) \textit{"Seattle"} is an implausible option. This underscores the challenge of ensuring consistency when manually constructing questions and answer choices. While CVQA is a valuable concurrent effort, our work focuses on a systematic approach that leverages VLM collaboration to efficiently build culturally specific datasets.

\begin{table}[t]
\centering
\small
\begin{tabular}{lcc}
\toprule
\textbf{Category} & \textbf{w/o Shot} & \textbf{w/ Shot} \\
\midrule
Food & 72.84 & 77.78 \\
Beverage & 76.32 & 76.32 \\
Game & 54.05 & 56.76 \\
Celebrations & 70.83 & 83.33 \\
Religion & 70.59 & 70.59 \\
Tool & 75.56 & 84.44 \\
Clothes & 81.82 & 86.36 \\
Heritage & 69.44 & 91.67 \\
Architecture & 82.50 & 87.50 \\
Agriculture & 67.50 & 85.00 \\
\midrule
\textbf{Total} & \textbf{74.31} & \textbf{81.30} \\
\bottomrule
\end{tabular}
\caption{Type 2 accuracy with and without one-shot prompting using corresponding Type 1 QA pairs. \texttt{gpt-4o-mini} is used for the experiments.}
\label{tab:type1-shot}
\end{table}

\subsection{Impact of Type 1 Grounding on Type 2 Reasoning}
\label{sec:app:type1-shot}
In Table \ref{tab:main_table}, we observe that models tend to perform better on Type 2 questions than on Type 1. This may seem counterintuitive, as Type 1 questions are grounded directly in visual cues. To examine whether grounding in Type 1 content can support Type 2 reasoning, we conducted an experiment where each Type 2 prompt was preceded by its corresponding Type 1 QA pair as a one-shot example. As shown in Table~\ref{tab:type1-shot}, this setup led to consistent gains across categories, improving overall accuracy from 74.31\% to 81.30\%. These results suggest visual grounding can meaningfully support higher-level cultural reasoning when leveraged appropriately.

\begin{table}[t]
\centering
\small
\begin{tabular}{lc}
\toprule
\textbf{Core Concept} & \textbf{IDS Chapter No.} \\
\midrule
Food & Food and Beverages (5) \\
Beverage & Food and Beverages (5) \\
Game & Motion (10) \\
Celebrations & Time (14), Religion and belief (22) \\
Religion & Religion and Belief (19) \\
Tool & Basic Actions and Technology (9) \\
Clothes & Clothing and Grooming (6) \\
Heritage & Cognition (17) \\
Architecture & The House (7) \\
Agriculture & Agriculture and Vegetation (8) \\
\bottomrule
\end{tabular}
\caption{Mapping between core concepts and IDS chapters. Some concepts share the same chapter due to overlapping semantic coverage.}
\label{tab:concept-ids-mapping}
\end{table}

\section{Dataset Construction Details}
\label{sec:app_data_detail}

\subsection{Mapping Core Concepts to IDS Chapters} 
\label{sec:app:ids_mapping}
As described in Section \ref{subsec:data_framework}, following \citet{marvl}, we define the concepts of our dataset based on the Intercontinental Dictionary Series (IDS) \citep{key_comrie_2015}. Table~\ref{tab:concept-ids-mapping} shows how our 10 core concepts map to chapters in the Intercontinental Dictionary Series (IDS). While some mappings are direct, others reflect one-to-many relationships based on semantic grouping and visual realizability (e.g., \textit{Food} and \textit{Beverage} both derive from Chapter 5). This mapping served as the basis for selecting culturally salient and visually grounded concepts in our dataset.

\subsection{Guidelines to \texttt{GPT-4-Turbo} for Dataset Annotation}
The detailed prompts for the annotation with GPT-4-Turbo are presented in Table~\ref{tab:type1_prompt} and Table~\ref{tab:type2_prompt}. 

\begin{table}[b!]
\begin{minipage}{0.99\columnwidth}\vspace{0mm} 
\begin{tcolorbox}
    \small
     \hspace{-3mm}
    \begin{tabular}{p{0.99\columnwidth}}

\begin{minipage}{0.99\columnwidth}\vspace{0mm}

{\bf Prompt for \textsc{Type 1} annotations: \\ \\}
\texttt{\textbf{[System Prompt]}}\\
You are a helpful Korean annotator to make visual question answering datasets.
\\

\texttt{\textbf{[User Prompt]}}\\
Given an image, generate a question asking for the name of the main object or the main activity that people are engaged in, and generate one correct option (answer) and three wrong options (distractors).\\

\textit{\textbf{Detailed guidelines are as follows:}}

1. The objects shown in the image is called \texttt{\textbf{``\{object\_name\}''}} in Korean. You can include this word into your correct options after translation into English. \\
2. All options should be written in up to 5 words.\\
3. Please struggle to make creative or challenging distractors so that they are not easily distinguished from the answer.\\
4. Distractors should seem similar to the correct answer and related to the category of the main object in image (e.g., Hanok - Agungi, Sarangchae, Anchae, Daecheongmaru). \\Distractors are better when they have similar color, shape, or texture with the answer.\\
5. Separate each distractor with ``;'' symbol.\\
6. Don't make any explanation.\\
7. Distractors should be culturally related to the image.\\
8. All the options should be either transliterated or translated. Never mix transliteration and translation.\\
9. Don't be too specific (Avoid using a proper name: instead use \texttt{[University building]} instead of \texttt{[Ewha Campus Complex]})\\
\\
\textit{\textbf{You can refer to below examples that are annotated for other images.}}\\
Question: What is the name of this place shown in the image?\\
Answer: Sarangchae\\
Distractors: Anchae ; Sadang ; Daecheongmaru\\

Question: What is the name of the structure seen in the image?\\
Answer: Ondol\\
Distractors: Agungi ; Jangdokdae ; Buttumak\\

Question: What is the name of this building shown in the image?\\
Answer: Gosiwon\\
Distractors: Officetel ; Apartment ; Share house\\

Please make four options (single answer and three distractors).

\end{minipage}
\end{tabular}
\end{tcolorbox}
    
\vspace{-2mm}
\caption{A prompt of GPT-4-Turbo used in the annotation of \textsc{Type 1} questions (i.e., \textit{visual recognition}) in the \textsc{Architecture} category.}
\label{tab:type1_prompt}
\end{minipage}
\end{table}
\begin{table}[b!]
\begin{minipage}{0.99\columnwidth}\vspace{0mm} 
\begin{tcolorbox}
    \small
     \hspace{-3mm}
    \begin{tabular}{p{0.99\columnwidth}}

\begin{minipage}{0.99\columnwidth}\vspace{0mm}

{\bf Prompt for \textsc{Type 2} annotations: \\ \\}
\texttt{\textbf{[System Prompt]}}\\
You are a helpful Korean annotator to make visual question answering datasets.\\

\texttt{\textbf{[User Prompt]}}\\
Please ask 5 questions and their options about the image. Here are the guidelines to follow for writing.\\

\textit{\textbf{Detailed guidelines are as follows:}}\\
1. The objects shown in the image is \texttt{\textbf{``\{object\_name\}''}}. Don't include this word in your questions.\\
2. The question should require looking at the image to answer.\\
3. Questions should require some knowledge about Korean cultures.\\ 
4. Don’t make a simple question that does not require knowledge of Korean cultures, such as recognizing objects or counting objects.\\
5. It is desirable to generate questions that are difficult for foreigners who are unfamiliar with Korean culture.\\
6. After writing a question, please write a single correct option (answer) and three wrong options (distractors) for your above question.\\
7. All options should be written in up to 5 words. \\
8. Don't ask traditional celebrations about the given image.\\ 
9. Try to ask questions that are more derived from the given image.\\
10. Any creative questions are very welcome.\\
11. Separate each distractor with ``;'' symbol.\\
12. [Description]\\
\textbf{\texttt{``\{object\_name\}\textbackslash n\{description\}''}\\}

\textbf{\textit{You can refer to below examples that are annotated for other images.}}

Question: Seen in the image, what traditional Korean heated floor system is associated with the heat source from this feature?\\
Answer: Ondol\\
Distractors: Daecheongmaru ; Anchae ; Buttumak\\

Question: In the image, what kind of Korean roof finishing is visible, known for its multicolored patterns?\\
Answer: Dancheong\\
Distractors: Seoggarae ; Cheoma ; Maru\\

Question: Which mode of transportation is commonly used by tourists to ascend the mountain where the tower is located?\\
Answer: Cable Car\\
Distractors: Bus ; Bicycle ; Funicular Railway\\

Please make four options (single answer and three distractors).

\end{minipage}
\end{tabular}
\end{tcolorbox}
    
\vspace{-2mm}
\caption{A prompt of GPT-4-Turbo used in the annotation of \textsc{Type 2} questions (i.e., \textit{cultural knowledge application}) in the \textsc{Architecture} category.}
\label{tab:type2_prompt}
\end{minipage}
\end{table}


\section{Dataset Analyses Details}
\subsection{Required Knowledge Analysis}
\label{sec:app_require_knowledge_detail}
We analyzed how diverse the cultural knowledge required by the questions in our \texttt{K-Viscuit} dataset is. To this end, we used the following prompt to obtain responses from the \texttt{GPT-4} model. We delivered \textit{all} \textsc{Type 2} samples (including both the questions and the options) to the model by concatenating them into a single string.


\begin{table}[h]
\begin{minipage}{0.99\columnwidth}\vspace{0mm} 
\begin{tcolorbox}
    \small
     \hspace{-3mm}
    \begin{tabular}{p{0.99\columnwidth}}

\begin{minipage}{0.99\columnwidth}\vspace{0mm}

{\bf Prompt for required knowledge analysis: \\ \\}
I created a multiple-choice quiz with four options per question, based on images related to Korean culture. Each question is designed to assess the understanding of one or more cultural elements. Please analyze which cultural element each question aims to measure and provide an overall summary.
\textbf{\texttt{``\{TYPE 2 samples\}''}}

\end{minipage}
\end{tabular}
\end{tcolorbox}
    
\vspace{-2mm}
\caption{Prompt for required knowledge analysis.}
\label{tab:prompts}
\end{minipage}
\end{table}

\subsection{Human Evaluation Details}
\label{sec:app_user_study}
We randomly selected images according to the proportion of each category to create the questionnaire for the human evaluation. If there were multiple \textsc{Type 2} questions for a single image, we sampled them randomly.
The number of selected images per category is as follows: \textsc{Food} (4), \textsc{Beverage} (2), \textsc{Game} (2), \textsc{Celebrations} (2), \textsc{Religion} (2), \textsc{Tool} (3), \textsc{Clothes} (2), \textsc{Heritage} (2), \textsc{Architecture} (4), and \textsc{Agriculture} (2). 

We released the survey on the Amazon MTurk platform, where non-Koreans with a relatively limited understanding of Korean culture were asked to complete the \texttt{K-Viscuit} questions within 20 minutes for a compensation of \$5.
The survey on the MTurk platform resulted in a demographic composed entirely of Americans.
Their self-assessed proficiency levels were: \textit{Very familiar} (35.7\%), \textit{Somewhat familiar} (50\%), \textit{Slightly familiar} (14.3\%), and \textit{Not familiar at all (0\%)}.
For Koreans, we administered the survey to 20 graduate students in their mid-to-late twenties.
We received feedback that Koreans had the most difficulty with questions related to history.

\section{VLM Evaluation Details}
\label{sec:app_model_details}
\paragraph{Model Implementation Details}
We present further implementation details in VLMs used in our experiments.
All open-source VLMs are implemented with the Transformers framework~\citep{wolf-etal-2020-transformers}, and the checkpoints are downloaded from Huggingface Hub\footnote{\url{https://huggingface.co/models}}.
For proprietary models, \texttt{gpt-4-turbo-2024-04-09}, \texttt{gemini-1.5-pro}, and \texttt{claude-3-opus} \texttt{-20240229} are used. The text prompt used for proprietary models is presented in Table~\ref{tab:closed_model_prompt}.

\begin{table}[ht!]
\begin{minipage}{0.99\columnwidth}\vspace{0mm} 
\begin{tcolorbox}
    \small
     \hspace{-3mm}
    \begin{tabular}{p{0.99\columnwidth}}

\begin{minipage}{0.99\columnwidth}\vspace{0mm}

{\bf Inference prompt for proprietary VLMs: \\ }

\texttt{\textbf{[System Prompt]}}\\
You will be given an image taken in Korea and a 4-way multiple-choice question. Answer the question based on the given image and your knowledge about Korean culture.\\

\texttt{\textbf{[User Prompt]}}\\
Question: \textbf{\texttt{``\{question\}''}}\\
Options:\\
a. \textbf{\texttt{``\{option\_a\}''}}\\
b. \textbf{\texttt{``\{option\_b\}''}}\\
c. \textbf{\texttt{``\{option\_c\}''}}\\
d. \textbf{\texttt{``\{option\_d\}''}}\\
\\
Make sure to respond with the option's letter: `a.', `b.', `c.', or `d.'.
Do not make any additional explanation.

\end{minipage}
\end{tabular}
\end{tcolorbox}
    
\vspace{-2mm}
\caption{Inference prompt for proprietary VLMs.}
\label{tab:closed_model_prompt}
\end{minipage}
\end{table}

\section{Retrieval Methodology Details}
\label{sec:app_retrieval}
For external knowledge retrieval, we generated image captions using \texttt{GPT-4} with the prompt shown in Table~\ref{tab:prompts_retrieval}.


\begin{table}[ht!]
\begin{minipage}{0.99\columnwidth}\vspace{0mm} 
\begin{tcolorbox}
    \small
     \hspace{-3mm}
    \begin{tabular}{p{0.99\columnwidth}}

\begin{minipage}{0.99\columnwidth}\vspace{0mm}

{\bf Prompt for generating image captions: \\ \\}
You are an AI language model specializing in identifying and describing Korean food from photographs. 
When given a photograph of Korean food, your task is to accurately describe the food based on its visual characteristics and visible ingredients. 
Your description should include the name of the dish, main ingredients, common accompaniments, and notable features that help identify the food. Be detailed yet concise, providing clear and helpful information to those trying to understand Korean cuisine. 
Ensure that your description is within 150 words.

\end{minipage}
\end{tabular}
\end{tcolorbox}
    
\vspace{-2mm}
\caption{Prompt for generating image captions.}
\label{tab:prompts_retrieval}
\end{minipage}
\end{table}

\end{document}